\documentclass[sigconf]{acmart}
\setcitestyle{numbers,sort&compress}

\newif\ifarxiv
\arxivtrue
\newif\ifanonymized
\anonymizedfalse

\newcommand{\showCODEN}[1]{}
\newcommand{\showDOI}[1]{}
\newcommand{\showISSN}[1]{}
\newcommand{\showLCCN}[1]{}
\newcommand{\showISBNx}[1]{}
\newcommand{\showISBNxiii}[1]{}
\newcommand{\shownote}[1]{}

\usepackage{booktabs}
\usepackage{amssymb,amsmath}
\usepackage[mathscr]{euscript} 
\usepackage{bbm}
\usepackage{dsfont}
\usepackage{mathtools}
\DeclarePairedDelimiter\abs{\lvert}{\rvert}

\newcommand{\mx}[1]{\mathbf{#1}}
\usepackage{booktabs}
\usepackage{siunitx}
\newcommand{\us}[1]{\SI{#1}{\micro\second}}
\newcommand{\ms}[1]{\SI{#1}{\milli\second}}
\newcommand{\fps}[1]{#1\,frame/s}
\usepackage[nomessages]{fp}
\newcommand{\calc}[2]{\FPeval{\calcresult}{round(#2,#1)}\calcresult}

\usepackage{subcaption}
\usepackage{multirow}
\newcolumntype{H}{>{\setbox0=\hbox\bgroup}c<{\egroup}@{}} 
\usepackage{pgfplots}
\usepackage{pgfplotstable}
\usepgfplotslibrary{groupplots}
\pgfplotsset{compat=newest}
\newcommand{\figref}[1]{Figure~\ref{#1}} 
\newcommand{\tblref}[1]{Table~\ref{#1}} 
\graphicspath{{./figs/},{./}} 

    \newcommand{\plotdataDir}[1]{./plotdata-merged/#1}

\newcommand{\pfmName}{Nvidia Tegra~X1}
\newcommand{\pfmVal}[2]{#2}

    \newcommand{\plotdataDirPfm}[1]{./plotdata-merged/#1}

\pgfplotscreateplotcyclelist{seqColorList}{
    blue!50!white,mark=diamond\\
    brown!70!black,mark=otimes\\
    red!90!black,mark=square\\
    orange,mark=star\\
    blue!80!black,mark=x\\
    green!80!black,mark=o\\
}

\ifarxiv
    \setcopyright{none}
\else
    \setcopyright{rightsretained}
\fi

    \renewcommand\footnotetextcopyrightpermission[1]{} 
    \acmConference{arXiv preprint}{v2}{June 2017}

\begin{document}

\title{CBinfer: Change-Based Inference for Convolutional Neural Networks on Video Data}

\ifanonymized
\else
    \author{Lukas Cavigelli}
    \orcid{0000-0003-1767-7715}
    \affiliation{\institution{ETH Zurich, Zurich, Switzerland}}
    \email{cavigelli@iis.ee.ethz.ch}
    \author{Philippe Degen}
    \affiliation{\institution{ETH Zurich, Zurich, Switzerland}}
    \email{degenp@ee.ethz.ch}
    \author{Luca Benini}
    \affiliation{\institution{ETH Zurich, Zurich, Switzerland}}
    \email{benini@iis.ee.ethz.ch}
    \renewcommand{\shortauthors}{L. Cavigelli et al.}
\fi

\addtolength{\abovecaptionskip}{-2mm}
\addtolength{\belowcaptionskip}{-1mm}

\begin{abstract}
Extracting per-frame features using convolutional neural networks for real-time processing of video data is currently mainly performed on powerful GPU-accelerated workstations and compute clusters. However, there are many applications such as smart surveillance cameras that require or would benefit from on-site processing.
To this end, we propose and evaluate a novel algorithm for change-based evaluation of CNNs for video data recorded with a static camera setting, exploiting the spatio-temporal sparsity of pixel changes. We achieve an average speed-up of $\pfmVal{7.6}{8.6}\times$ over a cuDNN baseline on a realistic benchmark with a negligible accuracy loss of less than 0.1\% and no retraining of the network. The resulting energy efficiency is $10\times$ higher than that of per-frame evaluation and reaches an equivalent of 328\,GOp/s/W on the Tegra~X1 platform. 
\end{abstract}

\maketitle


\section{Introduction}

Computer vision (CV) technology has become a key ingredient for automatized data analysis over a broad range of real-world applications: smart cameras for video surveillance, robotics, industrial quality assurance, medical diagnostics, and advanced driver assistance systems have recently become popular due the rising reliability of CV algorithms \cite{Porikli2013,Zhang2015b,Farrugia2009,Kaluarachchi2015}. This industry interest has fostered the procedure of a wealth of research projects yielding a fierce competition on many benchmarks datasets such as the ImageNet/ILSVRC \cite{Russakovsky2014,Deng2009}, MS COCO \cite{Lin2014}, and Cityscapes \cite{Cordts2016} benchmarks, on which scientists from academia and big industry players evaluate their latest algorithms. 

In recent years, the most competitive approaches to address many CV challenges have relied on machine learning with complex, multi-layered, trained feature extractors commonly referred to as deep learning \cite{He2015,Szegedy2014,Krizhevsky2012a}. The most frequently used flavor of deep learning techniques for CV are convolutional neural networks (ConvNets, CNNs). Since their landslide success at the 2012 ILSVRC competition over hand-crafted features, their accuracy has further improved year-over-year even exceeding human performance on this complex dataset \cite{HePReLU2015,Russakovsky2014}. CNNs keep on expanding to more areas of computer vision and data analytics in general \cite{Abu-El-Haija2016,HePReLU2015,Long2015,Kaluarachchi2015,Zhang2016a,Park2016}. 

Unfortunately, the high accuracy of CNNs comes with a high computational cost, requiring powerful GPU servers to train these networks for several weeks using hundreds of gigabytes of labeled data. While this effort is very costly, it is a one-time endeavour and can be done offline for many applications. However, the inference of state-of-the-art CNNs also requires several billions of multiplications and additions to classify even low resolution images by today's standards \cite{Cavigelli2015}. 
While in some cases offloading to centralized compute centers with powerful GPU servers is also possible for inference after deployment, it is extremely costly in terms of compute infrastructure and energy. Furthermore, collecting large amounts of data at a central site raises privacy concerns and the required high-bandwidth communication channel causes additional reliability problems and potentially prohibitive cost of deployment and during operation \cite{Kruegle1995}. 

The alternative, on-site near sensor embedded processing, largely solves the aforementioned issues by transmitting only the less sensitive, condensed information---potentially only security alerts in case of a smart surveillance camera---but imposes restrictions on available computation resources and power. 
These push the evaluation of such networks for real-time semantic segmentation or object detection out of reach of even the most powerful embedded platforms available today for high-resolution video data \cite{Cavigelli2015}. However, exactly such systems are required for a wide range of applications limited in cost (CCTV/urban surveillance, perimeter surveillance, consumer behavior and highway monitoring) and latency (aerospace and UAV monitoring and defence, visual authentication) \cite{Porikli2013,Kruegle1995}. 

Large efforts have thus already been taken to develop optimized software for heterogeneous platforms \cite{Chetlur2014,Cavigelli2015,Lavin2015,Lavin2015a,Vasilache2014,Jia2013}, to design specialized hardware architectures \cite{Cavigelli2016,Andri2016,Cavigelli2015a,Chen2016,Park2016,Farabet2011}, and to adapt the networks to avoid expensive arithmetic operations \cite{Rastegari2016,Courbariaux2015a,Zhang2016a}. However, they either do not provide a strong enough performance boost, are already at the theoretical limit of what can be achieved on a given platform, are inflexible and not commercially available, or incur a considerable accuracy loss. It is thus essential to extend the available options to efficiently perform inference on CNNs. 

In this paper, we propose a novel method to perform inference for convolutional neural networks on video data from a static camera with limited frame-to-frame changes. Evaluations on a \pfmName{}\footnote{The Nvidia Tegra X1 is a system-on-chip available on an embedded board with an affordable power budget (<15\,W) for a stationary camera.} show that an average speed-up of $\pfmVal{7.6}{8.6}\times$ is possible with negligible accuracy loss over cuDNN-based per-frame evaluation on an urban video surveillance dataset. This pushes real-time CNN inference on high-resolution frames within the computation and power budget of current embedded platforms. 

\paragraph*{Organization of the Paper} In the next section we will discuss related work, before proposing our change-based convolution algorithm in Section~\ref{sec:method}. We present experimental results and discuss them in in Section~\ref{sec:results}. We then conclude the paper in Section~\ref{sec:conclusion}.

\section{Related Work}
In this section, we will first discuss available datasets and CNNs with which we can evaluate our proposed algorithm. Then we describe existing optimized implementations for CNN inference and existing approximations trading accuracy for throughput. Finally, we survey related approaches exploiting the limited changes in video data to reduce the computational effort required to perform CNN inference. 

\subsection{Suitable Datasets and Neural Networks}
\begin{figure*}
	\centering 
	\includegraphics[width=0.95\linewidth]{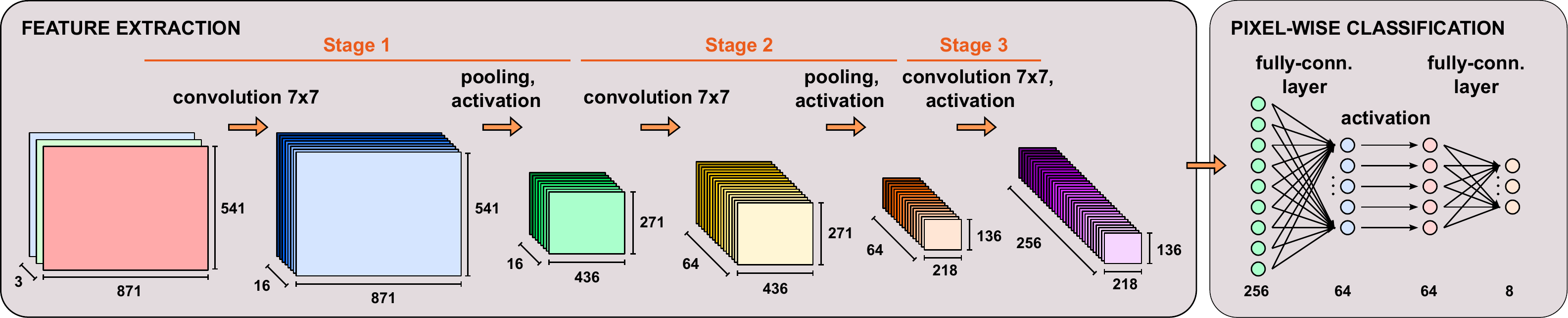}
	\caption{Schematic of the scene labeling convolutional neural network used for our evaluations \cite{Cavigelli2016a}.}
	\label{fig:convnet}
\end{figure*}
For our evaluations we are interested in performing object detection or semantic segmentation, which are both often applied to high-resolution images and video streams with frame rates above 10\,frame/s for meaningful applications. With still image object classification being considered solved by having achieved beyond human accuracy \cite{Russakovsky2014,He2015}, there is now a rapidly increasing interest in extracting information from video data, e.g. video tagging and action recognition on datasets that have recently become available (Sports-1M~\cite{Karpathy2014}, Youtube-8M~\cite{Abu-El-Haija2016}). 

\begin{figure}
	\centering
	\includegraphics[width=0.95\linewidth]{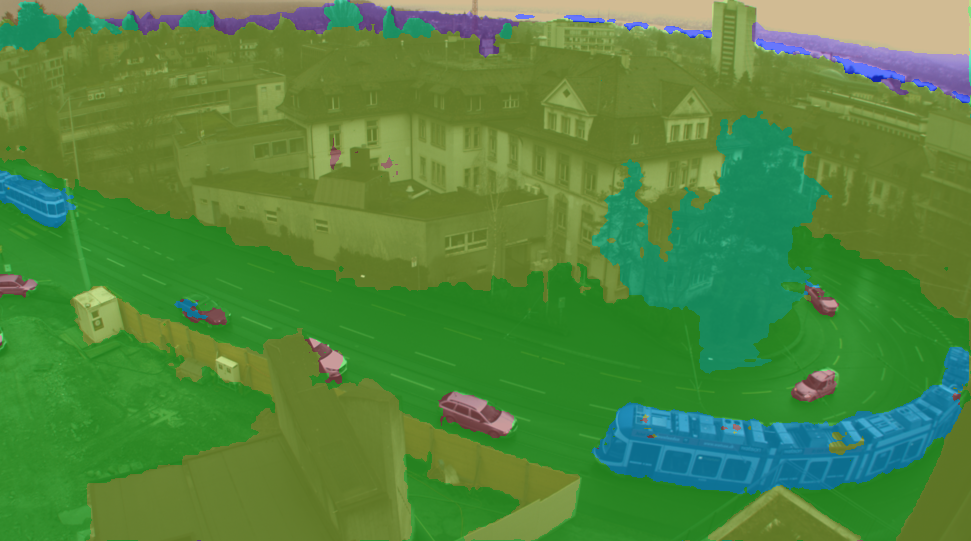}
	\caption{Example output of the scene labeling network of \cite{Cavigelli2016a} on which we evaluate our algorithm.}
	\label{fig:labelling}
\end{figure}

\begin{figure}
  \centering
  \includegraphics[width=0.95\linewidth]{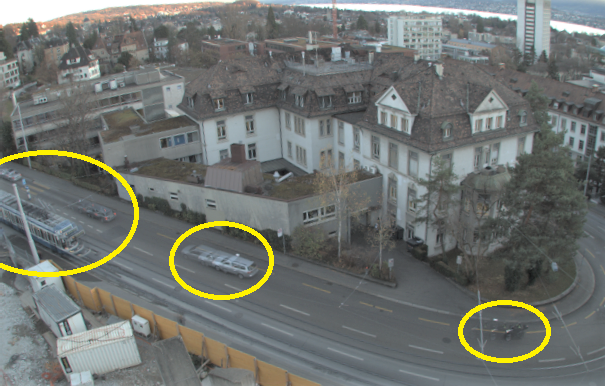}
  \caption{A sample video sequence from the dataset of \cite{Cavigelli2016a} showing the frame-by-frame changes by overlaying a sequence of length 3. Moving objects are only a small part of the overall scene and affect only a small share of the pixels.}
  \label{fig:gloria_seq}
\end{figure}

We are specifically interested in video sequences obtained from a static camera. While some such dataset exist, most of them are specifically targeted at person tracking and/or re-identification and do not provide labeled data for multi-class object detection or segmentation. However, the dataset used in \cite{Cavigelli2016a} provides ground truth labels for 10-class semantic segmentation from an urban street surveillance perspective, and while they work with individual images, several surrounding unlabeled frames and a trained convolutional network are available. An example image labeled with the provided CNN is shown in \figref{fig:labelling}, and a sample sequence of 3 images is visualized in \figref{fig:gloria_seq}.

\subsection{Optimized Embedded System Implementations} \label{sec:relWorkImpl}
The latest wave of interest in neural networks can be attributed to their sudden success driven by the availability of large datasets and the increasingly powerful computing platforms. One of the most economical and practicable solutions for training medium-sized CNNs is to use a workstation with GPUs. The available software frameworks to implement and train CNNs provide strong support for this kind of platform. 

The massive amounts of compute time spent training CNNs has spurred the development of highly optimized GPU implementations. First, most widely used frameworks relied on their own custom implementations which have all converged to methods relying on matrix-multiplications \cite{Jia2013,Collobert2011}, leveraging the availability of highly optimized code in BLAS libraries and the fact that GPUs are capable of achieving a throughput within a few percent of their peak performance with this type of workload. Specialized libraries such as Nvidia's cuDNN and Nervana Systems' Neon provide some additional performance gains through assembly-level implementations \cite{Lavin2015} and additional algorithmic improvements such as Winograd and FFT-based convolution \cite{Lavin2015a}. A specific implementation for non-batched inference on an embedded platform building on a matrix multiplication is documented in \cite{Cavigelli2015}, also showing that more than 90\% of time is spent computing convolutions.

\subsection{Approximations Trading Accuracy for Throughput} \label{sec:approxCNN}
Admitting limited accuracy losses in order to gain a higher throughput by approximating existing networks, inference algorithms, and arithmetic operations can help overcome the computational obstacles preventing widespread adoption of CNN-based algorithms on embedded and mobile platforms. 

One such option is the reduction of the required arithmetic precision to evaluation NNs. Various methods from normal fixed-point analysis to retraining networks to adapt for quantized weights and activations exist. While most fixed-point methods are of limited use on many off-the-shelf software programmable platforms, some can benefit from vectorization of lower-precision operations \cite{Gysel2016a}. Extreme methods go as far as to enforce binary weights \cite{Courbariaux2015a,Andri2016}, and in some cases also binary activations \cite{Rastegari2016}. This means that multiplications can be dropped entirely, and in case of binary activations even collapse some of the add/subtract operations into XNOR and bit count operations. Many networks can be quantized with 8\,bit without an increase in error rate, before there is a trade-off between precision and accuracy \cite{Cavigelli2016,Hashemi2016}. Some methods try reducing the computational effort by pruning many very small weights to zero, making it possible to skip some operations \cite{Li2016}. More sophisticated quantization schemes such as vector quantization exist and can further compress a trained CNN model, but they require specialized hardware to bring an improvement in energy efficiency \cite{Han2016a,Aimar2017}. Once focusing on application-specific accelerators, also approximate arithmetic such as inaccurate multipliers have been considered \cite{Zhang2015a}. 

Further research has focused on optimizing semantic segmentation and object detection algorithms to better reuse already computed features by eliminating any non-convolutional elements from the network \cite{Redmon2016,Ren2015,Long2015}. Simplifying the operations in a network, such as low-rank approximations of 2D convolutions or by simply designing smaller networks with state-of-the-art methods have been evaluated in \cite{Jaderberg2014a,Iandola2016,Paszke2016}. 

The method we propose in this paper does not supersede these methods, but can be combined with the aforementioned approximation methods to further improve throughput.

\subsection{Video-based Computation Reduction}
Obtaining per-frame features naturally seems like an easier task when these frames belong to a video sequence rather than a random collection of images. Limited movement of objects in a frame can be exploited in object tracking by working with a limited search window within the frame \cite{Held2016}, not only reducing the problem size, but also simplifying the regression task---up until the tracked target is occluded by a large object.

For object detection and semantic segmentation, the available work in this direction is limited to clockwork CNNs \cite{Shelhamer2016}. The authors of \cite{Long2015} have extended their work on fully convolutional networks for semantic segmentation, which presents a CNN with skip connections and deconvolution layers to refine the lower-resolution feature maps obtained deep within the network using the features extracted early in the network. They exploit the fact that lower-resolution feature maps within the network are more stable over time than the full-resolution input. They thus propose to reevaluate the first few layers and the last few affected through the skip connections more frequently than the more coarse grained feature maps. This is a strong limitation on the set of CNNs this method can be applied to. They present evaluations based on a static as well as a dynamic, content-adaptive reevaluation schedule, showing that they can reduce the number of full-frame convolutions by about 40\% before the accuracy starts to drop on the Youtube-Objects dataset. 

However, this approach is limited to updating entire frames, whereas we exploit that often only small parts of the scene change and need to be reevaluated. We are not aware of any existing methods exploiting limited changes between frames, which we show to allow for much larger gains in throughput.

\section{Methodology} \label{sec:method}
Differently from to previous work looking at reevaluating entire frames, we exploit the limited number of pixels changing frame-to-frame to increase the throughput without loss in classification accuracy. 
The most straight-forward pixel-level approach is to detect changing pixels at the input based on a threshold on the difference to the previous frame and then update all the pixels affected by them, increasing the number of pixels to be updated layer-after-layer due to the convolution operations. Thus for e.g. a $7\times 7$ convolution a one-pixel change triggers an update of 49 pixels in the next layer and 169 pixels after another $7\times 7$ convolution. Strided operations (often used with pooling layers) reduce this effect, but do not prevent it. This issue might seem prohibitive for multi-layer CNNs, particularly when considering that individual pixels might keep exceeding the threshold due to noise. 

However, the change is not only spatially local at the input, but also at the output. Furthermore, noise-like changes will likely not have strong impacts on feature maps deeper within the network. We thus propose to perform the change-detection not only at the input, but before each convolution layer---relative to its previous input---and to compute an updated value only for the affected output pixels. This can be done without modifications to the training of the CNN, can be applied to existing pre-trained networks, and is not specific to the CNN on which we evaluate the proposed algorithm. 

We propose to replace all spatial convolution layers (conv layers) with \emph{change-based} spatial convolution layers (CBconv layers). This means adapting the widely used, simple and well-performing matrix-generation and matrix-multiplication sequence of operations \cite{Jia2013,Cavigelli2015}. The convolution layer computes
\begin{small}
\begin{align}
    y_o(j,i) = b_o + \sum_{c\in\mathcal{C}_{in}}\sum_{(\Delta j,\Delta i)\in\mathcal{S}_k} k_{o,c}(\Delta j,\Delta i) x_c(j-\Delta j,i-\Delta i),
\end{align}
\end{small}
where $o$ indexes the output channels $\mathcal{C}_{out}$ and $c$ indexes the input channels $\mathcal{C}_{in}$. The pixel is identified by the tuple $(j,i)$ and $\mathcal{S}_k$ denotes the support of the filters kernels $k$. 
This can be computed by performing a matrix multiplication
\begin{align}
	\mx{Y} = \mx{K} \mx{X}&, 
	\quad \mx{Y}\in\mathbb{R}^{|\mathcal{C}_O| \times h_o \cdot w_o}, \label{eq:matMult}\\
	\quad \mx{K}\in\mathbb{R}^{|\mathcal{C}_O| \times |\mathcal{C}_I| \cdot h_k \cdot w_k}&, 
	\quad \mx{X}\in\mathbb{R}^{|\mathcal{C}_I| \cdot h_k \cdot w_k \times h_o \cdot w_o}.
\end{align}
The image matrix $\mx{X}$ is constructed as $X((k h_k + j)w_k + i, y_o w_o + x_o) = x(k, j+y_o, i+x_o)$ with $k=1,\dots,|\mathcal{C}_{in}|$, $j=1,\dots,h_k$, $i=1,\dots,w_k$ and $y_o=1,\dots,h_o$, $x_o=1,\dots,w_o$. 
The filter matrix $\mx{K}$ is given by $K(o, (c h_k + j) w_k + i) = k(o,c,j,i)$ for $o=1,\dots,|\mathcal{C}_{out}|$, $c=1,\dots,|\mathcal{C}_{in}|$, $j=1,\dots,h_k$ and $i=1,\dots,w_k$. 
The result matrix is stored as $Y(o, y_o w_o + x_o) = y(o, y_o, x_o)$. Zero-padding can be applied during the construction of the $\mx{X}$ matrix and an efficient strided convolution can be computed by dropping the unused rows. 

We replace this matrix multiplication by the following sequence of processing steps, thereby drastically reducing the size of the matrix used in the main computation step. 

\subsection{Processing Steps}
\begin{figure*}
  \centering
  \includegraphics[width=\linewidth]{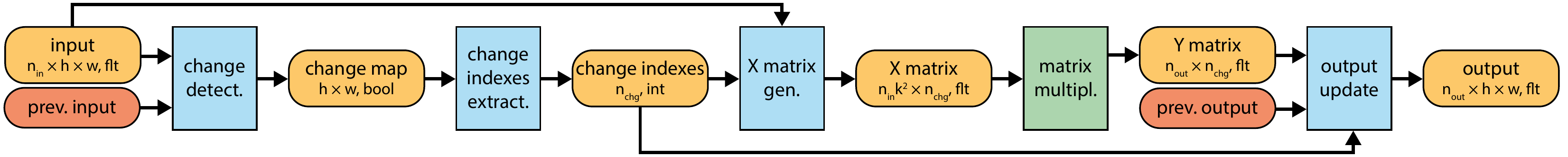}
  \caption{Processing flow of the change-based convolution algorithm. Custom processing kernels are shown in blue, processing steps using available libraries are shown in green, variables sharable among layers are shown in yellow, and variables to be stored per-layer are colored orange. The size and data type of the tensor storing the intermediate results is indicated below each the variable name.}
  \label{fig:algostruct}
\end{figure*}
We modify the standard approach and use a sequence of processing steps (cf. \figref{fig:algostruct}): change detection, change indexes extraction, matrix generation, matrix multiplication, and output update. In the following, we will explain the individual steps. 
\paragraph{Change Detection}
In this step, changed pixels are detected. We define a changed pixel as one where the absolute difference of the current to the previous input of any feature map/channel exceeds some threshold $\tau$, i.e.
\begin{align*}
    m(j,i) = \bigvee_{c\in\mathcal{C}_I}\abs{x^{(t)}(c,j,i)-x^{(t-1)}(c,j,i)}>\tau.
\end{align*}
The computation effort of this step is crucial, since it is executed independently of whether any pixel changed. Each of these changes affects a region equal to the filter size, and these output pixels are marked for updating: 
\begin{align*}
    \widetilde{m}(j,i) = \bigvee_{(\Delta j,\Delta i)\in\mathcal{S}_k} m(j+\Delta j,i+\Delta i),  
\end{align*}
where $\mathcal{S}_k$ is the filter kernel support, e.g. $\mathcal{S}_k=\{-3,\dots,3\}^2$ for a $7\times 7$ filter. 
All of this is implemented on GPU by clearing the the change map to all-zero and having one thread per pixel, which---if a change is detected---sets the pixels of the filter support neighborhood in the resulting \emph{change map}. 

\paragraph{Change Indexes Extraction}
In this step, we condense the change map $\widetilde{m}$ to 1) a list of pixel indexes where changes occurred and 2) count the number of changed pixels. This cannot easily be performed in parallel, so for our implementation we split the change map into blocks of pixels, compute the result for all the blocks in parallel, and reassemble the result. The computed index list is later on needed to access the right pixels to assemble the matrix for the convolution. 

\paragraph{Matrix Generation \& Matrix Multiplication}
Matrix multiplications are used in many applications, and highly optimized implementations such as the GEMM (general matrix multiplication) function provided by the Nvidia cuBLAS library come within a few percent of the peak FLOPS of which a GPU is capable to provide. Matrix multiplication-based implementations of the convolution layer relying on this are widely available and are highly efficient \cite{Jin2014,Cavigelli2015} and is described earlier in this section. The $\mx{X}$ matrix in \eqref{eq:matMult} is not generated full-sized, but instead only those columns corresponding to the relevant output pixels are assembled, resulting in a reduced width equal to the number of output pixels affected by the changes in the input image. The $\mx{K}$ matrix is made up of the filters trained using normal convolution layers and keeps the same dimensions, so the computation effort in this step is proportional to the number of changed pixels and the matrix multiplication is in the worst case only as time consuming as the full-frame convolution. 

\paragraph{Output Updating}
We use the previously stored results and the newly computed output values along with the change indexes list to provide the updated output feature maps. To maximize throughput, we also include the ReLU activation of the affected pixels in this step.

\subsection{Memory Requirements}
The memory requirements of DNN frameworks are known to be very high, up to the point where it becomes a limiting factor for increasing the mini-batch size during learning and thus reducing the throughput when parallelizing across multiple GPUs. These requirements are very different when looking at embedded inference-only systems: 
\begin{enumerate}
    \item Inference is typically done on single frames and creating mini-batches would introduce often unacceptable latency and the benefit of doing so is limited to a few percent of additional performance \cite{Cavigelli2015}. 
    \item To maximize modularity and because it is required during training, each layer typically has memory allocated to store its output with the exception of ReLU activation layers which are often applied in-place. 
    \item To keep a high modularity, the memory to keep the matrix $\mx{X}$ is often not shared among layers, although its values are never reused after finishing the convolution computation. 
    \item Batch normalization layers (if present) are considered independent layers with their own output buffer, but they can be absorbed into the convolution layer for inference. 
\end{enumerate}
To obtain a baseline memory requirement, we compute the required memory of common DNN frameworks performing convolutions using matrix multiplication with a batch size of 1. We assume an optimized network minimizing the number of layers, e.g. by absorbing batch normalization layers into the convolution layers or using in-place activation layers. This way 30M values need to be stored for the intermediate results, 264M values for the $\mx{X}$ matrix, and 873k values for the parameters. This can further be optimized by sharing  $\mx{X}$ among all convolution layers and by keeping only memory allocated to storing only the output of two layers and switching back-and-forth between them, layer-by-layer. This reduces the memory footprint to 9M, 93M, and 872k values, and a total of 103M values for our baseline.

Applying our algorithm requires a little more memory, because we need to store additional intermediate results (cf. \figref{fig:algostruct}) such as the change matrix, the changed indexes list, and the $\mx{Y}$ matrix, which can all again be shared between the layers. We also need to store the previous output to use it as a basis for the updated output and to use it as the previous input of the subsequent layer. For our sample network, this required another $\sim$\,60M values to a total of 163M values (+\calc{0}{100/103*(103+60)-100}\%, total size $\sim$\,650\,MB)---an acceptable increase and not a limitation, considering that modern graphics cards typically come with 8\,GB memory and even GPU-accelerated embedded platforms such as the Nvidia Jetson~TX1 module provide 4\,GB of memory.

\subsection{Threshold Selection} \label{sec:thSelectionProc}
The proposed algorithm adds one parameter to each convolution layer, the detection threshold. It is fixed offline after the training based on sample video sequences. A threshold of zero should yield identical results to the non-change-based implementation, which has been used for functional verification. For our evaluations we used the following procedure to select the thresholds: We start by setting all thresholds to zero. Then we iteratively step through them from the first to the last layer, sweeping the threshold parameter for each layer and keeping the maximum value before a clear performance degradation became noticeable when evaluating the entire validation set. The following evaluations will show that these thresholds need not be re-calibrated per video sequence and neither the accuracy nor the speed-up are overly sensitive to them.

\section{Results \& Discussion} \label{sec:results}
In this section, we will first present the evaluation environment and analysis the baseline compute time breakdown. We then show how the threshold parameters have been selected before discussing throughput measurements and the accuracy-throughput trade-off. Finally, we discuss the compute time breakdown and how changes propagate through the network to confirm the quality of our GPU implementation and justify design choices made during the construction of the algorithm. 

\subsection{Evaluation Environment}
\begin{figure}
  \centering
  \includegraphics[width=0.95\linewidth]{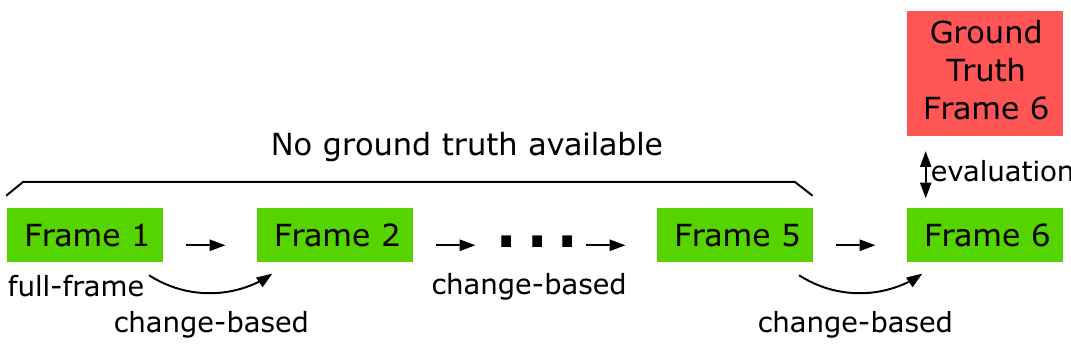}
  \caption{Scheme of the image sequence used for the evaluations.}
  \label{fig:seqSchema}
\end{figure}
While the algorithm is not limited to scene labeling/semantic segmentation, we perform our evaluations on the urban surveillance dataset described in \cite{Cavigelli2016a} and using the corresponding scene labeling CNN, not using the multispectral imaging data. The dataset provides 51 training images and 6 validation images with $776\times 1040$ pixel with the corresponding ground-truth scene labeling, classifying each pixel into one of the following 8 classes: building, road, tree, sky, tram, car/truck, water, distant background. 
For the validation set, the labeled images are part of short video sequences with 5 additional frames available before the frame for which the ground truth labeling is available. 
A trained network on this data is described in \cite{Cavigelli2016a} and its parameters are reused unaltered for our evaluations. 
The procedure with which we perform our evaluation is visualized in \figref{fig:seqSchema}.

We have implemented the proposed algorithm using CUDA and wrapped them as modules for the Torch framework~\cite{Collobert2011}. 
\pfmVal{
The machine on which we evaluated the performance has an Intel Xeon E5-1620v2 CPU and a Nvidia GTX 1080 GPU. 
}{
We have evaluated the performance on a Jetson TX1 board with JetPack~2.3.1 (Nov. 2016). 
}
Our performance baseline for the entire CNN and for the change-based implementation the pixel-wise classification is relying on Nvidia's cuDNN \pfmVal{v5.1.10 (Nov. 2016)}{v5.1.5}, which includes optimizations such as the Winograd algorithm and FFT-based convolutions mentioned in Section~\ref{sec:relWorkImpl}.

\subsection{Baseline Throughput and Computation Breakdown} \label{sec:baselineComputeBreakdown}
\begin{table}
    \centering
    \caption{Performance Baseline Compute Time Breakdown}
    \label{tbl:baselineComputeTime}
    \begin{tabular}{c|rrr|rr}
        \toprule
         Layer & Conv. & Activ. & Pooling & total & share \\ \midrule
         1 & \pfmVal{\us{306}}{\ms{72.9}} & \pfmVal{\us{502}}{\ms{7.4}} & \pfmVal{\us{278}}{\ms{3.3}} & \pfmVal{\us{4086}}{\ms{83.6}} & \pfmVal{13.3}{15.6}\% \\
         2 & \pfmVal{\us{7379}}{\ms{116.2}} & \pfmVal{\us{510}}{\ms{6.9}} & \pfmVal{\us{322}}{\ms{3.3}} & \pfmVal{\us{8210}}{\ms{126.4}} & \pfmVal{26.7}{23.6}\% \\
         3 & \pfmVal{\us{16972}}{\ms{302.8}} & \pfmVal{\us{501}}{\ms{6.6}} & --- & \pfmVal{\us{17473}}{\ms{309.4}} & \pfmVal{56.7}{57.8}\% \\
         4 & \pfmVal{\us{773}}{\ms{12.7}} & \pfmVal{\us{12}}{\ms{1.7}} & --- & \pfmVal{\us{902}}{\ms{14.4}} & \pfmVal{2.9}{2.7}\% \\
         5 & \pfmVal{\us{125}}{\ms{1.6}} & --- & --- & \pfmVal{\us{125}}{\ms{1.6}} & \pfmVal{0.4}{0.3}\% \\ \bottomrule
    \end{tabular}

\end{table}
Before we discuss the performance of the proposed algorithm, we analyze the baseline throughput and compute time breakdown in \tblref{tbl:baselineComputeTime}. Clearly, most time is spent performing convolutions, and the layers 1--3 performing $7\times 7$ convolutions and belonging to the feature extraction part of the network are dominant with \pfmVal{96.7\%}{91.9\% (\ms{492})} of the overall computation time (\ms{\pfmVal{30.8}{535}} or \fps{\pfmVal{32.5}{1.87}}). We thus specifically focus our analyses on these 3 layers, replacing only them with our CBconv layer. 

\subsection{Threshold Selection} \label{sec:ThresholdSelResults}
\begin{figure*}
  \centering
    \begin{tikzpicture}
        \begin{groupplot}
            [
                group style = {
                    group size=3 by 1,
                    ylabels at=edge left,
                    yticklabels at=edge left,
                    horizontal sep=1mm,
                },
                ylabel={Error Incr. [\%]},
                ymin=-0.075, ymax=0.8,
                width=0.39\linewidth, 
                height=37mm,
                grid=major, 
                cycle list name=seqColorList,
            ]
            \nextgroupplot[xlabel={Threshold for Layer 1 ($\tau_1$)},]
            \addplot+ [] table 
                [x=threshold, y=errorDiff, col sep=comma]
                {\plotdataDir{outp-eval2-mode_thr1-pictureset15-cb.csv}};
            \nextgroupplot[xlabel={Threshold for Layer 2 ($\tau_2$)},]
            \addplot+ [] table [x=threshold, y=errorDiff, col sep=comma]
                {\plotdataDir{outp-eval2-mode_thr2-pictureset15-cb.csv}};
            \nextgroupplot[xlabel={Threshold for Layer 3 ($\tau_3$)},]
            \addplot+ [] table [x=threshold, y=errorDiff, col sep=comma]
                {\plotdataDir{outp-eval2-mode_thr3-pictureset15-cb.csv}};
        \end{groupplot}
    \end{tikzpicture}
  \caption{Analysis of the increase in pixel classification error rate by selecting a certain change detect threshold. This analysis is conducted layer-by-layer, where the error increase of any layer includes the error introduced by the previous layers' threshold choice ($\tau_1=0.04, \tau_2=0.3, \tau_3=1.0$).}
  \label{fig:thSelection}
\end{figure*}
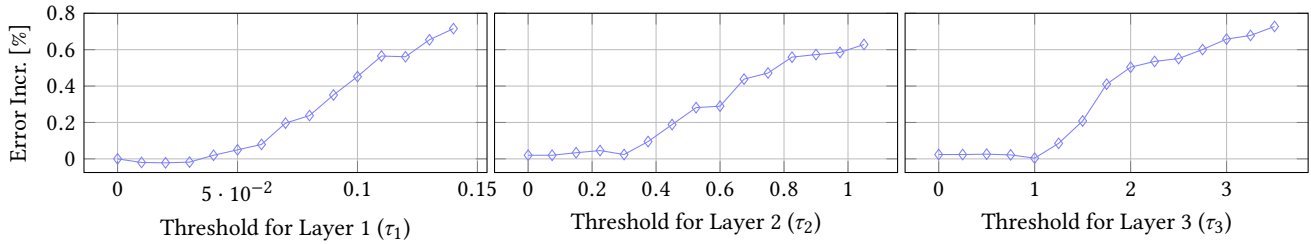
Our algorithm introduces a threshold parameter for each layer, for which we outline the selection process in Section~\ref{sec:thSelectionProc}. While we might want to leave them variable to investigate a throughput against accuracy trade-off, we also want to ensure that not a single layer's threshold is limiting the overall accuracy by aligning the tipping point where the accuracy starts to drop. We choose the thresholds conservatively, accepting very little accuracy drop, since any classification error will be focused around the moving objects which are our area of interest. We sweep the parameters of each layer to determine the increase in error (cf.~\figref{fig:thSelection}). We do so first for Layer~1 with $\tau_2=\tau_3=0$ and select $\tau_1=0.04$, before repeating it for layers 2 and 3 after each other and using the already chosen thresholds for the previous layers, selecting $\tau_2=0.3$ and $\tau_3=1.0$. 

With this selection of thresholds we can scale them jointly to analyze the trade-off against the classification accuracy more concisely (cf. \figref{fig:allEvals}, left). The accuracy of the individual test sequences are visualized, and clearly show a similar behavior with a plateau up to a clear point where there is a steep increase in error rate.

\subsection{Throughput Evaluations}
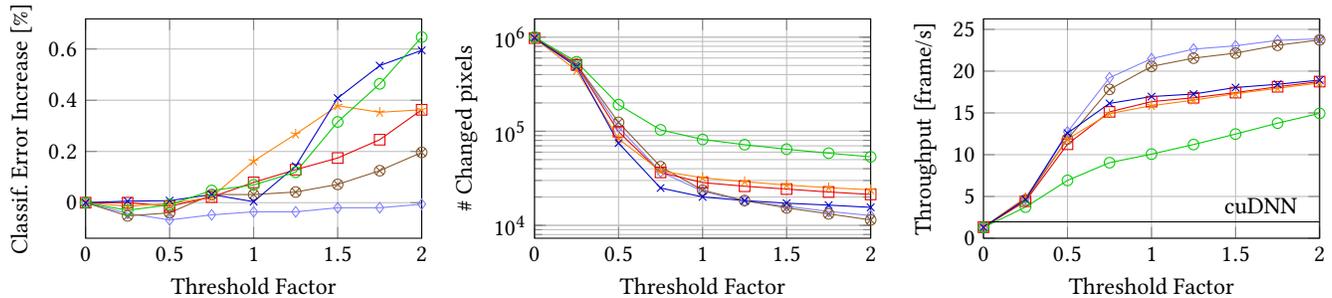
\begin{figure*}
  \centering
    \begin{tikzpicture} 
        \pgfplotstableread[col sep=comma]
            {\plotdataDirPfm{outp-eval2-mode_thrFactor-pictureset3-baseline.csv}}\dataBaseline;
        \pgfplotstablegetelem{0}{throughput}\of\dataBaseline
        \pgfmathsetmacro{\baselineVal}{\pgfplotsretval}

        \begin{groupplot}
            [
                group style = {
                    group size=3 by 1,
                    horizontal sep=15mm,
                },
                width=0.34\linewidth, 
                height=45mm,
                cycle list name=seqColorList,
                xmin=0, xmax=2, 
            ]
            \nextgroupplot[
                xlabel={Threshold Factor},
                ylabel={Classif. Error Increase [\%]}, 
                grid=major, 
                try min ticks=6, 
            ]
            \addplot+ table [x=threshold, y=errorDiff, col sep=comma]
                {\plotdataDirPfm{outp-eval2-mode_thrFactor-pictureset3-cb.csv}};
            \addplot+ table [x=threshold, y=errorDiff, col sep=comma]
                {\plotdataDirPfm{outp-eval2-mode_thrFactor-pictureset4-cb.csv}};
            \addplot+ table [x=threshold, y=errorDiff, col sep=comma]
                {\plotdataDirPfm{outp-eval2-mode_thrFactor-pictureset9-cb.csv}};
            \addplot+ table [x=threshold, y=errorDiff, col sep=comma]
                {\plotdataDirPfm{outp-eval2-mode_thrFactor-pictureset10-cb.csv}};
            \addplot+ table [x=threshold, y=errorDiff, col sep=comma]
                {\plotdataDirPfm{outp-eval2-mode_thrFactor-pictureset15-cb.csv}};
            \addplot+ table [x=threshold, y=errorDiff, col sep=comma]
                {\plotdataDirPfm{outp-eval2-mode_thrFactor-pictureset16-cb.csv}};
            \nextgroupplot[
                xlabel={Threshold Factor},
                ylabel={\# Changed pixels}, 
                grid=both, 
                try min ticks=6, 
                ymode=log,
            ]
            \addplot+ table [x=threshold, y=numChangeTotal, col sep=comma]
                {\plotdataDirPfm{outp-eval2-mode_thrFactor-pictureset3-cb.csv}};
            \addplot+ table [x=threshold, y=numChangeTotal, col sep=comma]
                {\plotdataDirPfm{outp-eval2-mode_thrFactor-pictureset4-cb.csv}};
            \addplot+ table [x=threshold, y=numChangeTotal, col sep=comma]
                {\plotdataDirPfm{outp-eval2-mode_thrFactor-pictureset9-cb.csv}};
            \addplot+ table [x=threshold, y=numChangeTotal, col sep=comma]
                {\plotdataDirPfm{outp-eval2-mode_thrFactor-pictureset10-cb.csv}};
            \addplot+ table [x=threshold, y=numChangeTotal, col sep=comma]
                {\plotdataDirPfm{outp-eval2-mode_thrFactor-pictureset15-cb.csv}};
            \addplot+ table [x=threshold, y=numChangeTotal, col sep=comma]
                {\plotdataDirPfm{outp-eval2-mode_thrFactor-pictureset16-cb.csv}};
            \nextgroupplot[
                xlabel={Threshold Factor},
                ylabel={Throughput [frame/s]}, 
                grid=major,
                try min ticks=6, 
                ymin=0,
            ]
            \addplot+ table [x=threshold, y=throughput, col sep=comma]
                {\plotdataDirPfm{outp-eval2-mode_thrFactor-pictureset3-cb.csv}};
            \addplot+ table [x=threshold, y=throughput, col sep=comma]
                {\plotdataDirPfm{outp-eval2-mode_thrFactor-pictureset4-cb.csv}};
            \addplot+ table [x=threshold, y=throughput, col sep=comma]
                {\plotdataDirPfm{outp-eval2-mode_thrFactor-pictureset9-cb.csv}};
            \addplot+ table [x=threshold, y=throughput, col sep=comma]
                {\plotdataDirPfm{outp-eval2-mode_thrFactor-pictureset10-cb.csv}};
            \addplot+ table [x=threshold, y=throughput, col sep=comma]
                {\plotdataDirPfm{outp-eval2-mode_thrFactor-pictureset15-cb.csv}};
            \addplot+ table [x=threshold, y=throughput, col sep=comma]
                {\plotdataDirPfm{outp-eval2-mode_thrFactor-pictureset16-cb.csv}}; \label{plt:thrghpt16}
            \draw[thin] (axis cs:\pgfkeysvalueof{/pgfplots/xmin},\baselineVal) -- (axis cs:\pgfkeysvalueof{/pgfplots/xmax},\baselineVal)
            node[above] at (axis cs:{\pgfkeysvalueof{/pgfplots/xmax}-0.35},\baselineVal) {cuDNN};
        \end{groupplot}
    \end{tikzpicture}
  \caption{Evaluation of the impact of jointly scaling the change detection thresholds on the classification error, the number of detected changed pixels (sum over all 3 layers), and the throughput.}
  \label{fig:allEvals}
\end{figure*}
The motivation for the entire proposed algorithm was to increase throughput by focusing only on the frame-to-frame changes. We show the performance gain in \figref{fig:allEvals} (right) with the indicated baseline analyzing the entire frame with the same network using cuDNN. In the extreme case of setting all thresholds to zero, the entire frame is updated, which results in a clear performance loss because of the change detection overhead as well as fewer optimization options such as less cache-friendly access patterns when generating the $\mx{X}$ matrix. 

When increasing the threshold factor, the throughput increases rapidly to about \pfmVal{200-250}{16}\,frame/s, where it starts saturating because the change detection step as well as other non-varying components like the pooling and pixel classification layers are becoming dominant and the number of detected changed pixels does not further decrease. We almost reach this plateau already for a threshold factor of 1, where we have by construction almost no accuracy loss. The average frame rate over the different sequences is near \pfmVal{200}{17}\,frame/s at this point---an improvement of $\pfmVal{7.6}{8.6}\times$ over the cuDNN baseline of \pfmVal{26.35}{1.96}\,frame/s. 

One sequence (\ref{plt:thrghpt16}) has---while still being close to $\pfmVal{3}{5.1}\times$ faster than the baseline---a significantly lower throughput than the other sequences. While most of them show typical scenarios such as shown in \figref{fig:gloria_seq}, this sequences shows a very busy situation where the entire road is full of vehicle and all of them are moving. The aggregate number of changed pixels across all 3 layers is visualized in \figref{fig:allEvals} (center). Most sequences trigger less than 3\% of the maximum possible number of changes while the aforementioned exceptional case has a significantly higher share of around 9\%. 

We have repeated the same evaluations on a workstation with a Nvidia GTX Titan X GPU, obtaining an almost identical throughput-threshold trade-off and compute time breakdown up to a scaling factor of ~$11.9\times$---as can be expected for a largely very well parallelizable workload and a $12\times$ more powerful device with a similar architecture (TX1: 512 GFLOPS and 25.6 GB/s DRAM bandwidth, GTX Titan X: 6144 GFLOPS and 336 GB/s).

\subsection{Accuracy-Throughput Trade-Off}\begin{figure}
  \centering
      \begin{tikzpicture}
    
        \pgfplotstableread[col sep=comma]
            {\plotdataDirPfm{outp-eval2-mode_thrFactor-pictureset16-baseline.csv}}\dataBaseline;
        \pgfplotstablegetelem{0}{throughput}\of\dataBaseline
        \pgfmathsetmacro{\baselineVal}{\pgfplotsretval}
        
        \begin{axis}[
                width=1.0\linewidth,
                height=0.65\linewidth, 
                xlabel={Pixel Classification Accuracy [\%]},
                ylabel={Throughput [frame/s]}, 
                xmin=91, xmax=99, ymin=0, grid=major, 
                try min ticks=6, 
                cycle list name=seqColorList,
            ]
        
            \addplot+ table 
                [x=accPixel, y=throughput, col sep=comma]
                {\plotdataDirPfm{outp-eval2-mode_thrFactor-pictureset3-cb.csv}};
            \addplot+ table 
                [x=accPixel, y=throughput, col sep=comma]
                {\plotdataDirPfm{outp-eval2-mode_thrFactor-pictureset4-cb.csv}};
            \addplot+ table 
                [x=accPixel, y=throughput, col sep=comma]
                {\plotdataDirPfm{outp-eval2-mode_thrFactor-pictureset9-cb.csv}};
            \addplot+ table 
                [x=accPixel, y=throughput, col sep=comma]
                {\plotdataDirPfm{outp-eval2-mode_thrFactor-pictureset10-cb.csv}};
            \addplot+ table 
                [x=accPixel, y=throughput, col sep=comma]
                {\plotdataDirPfm{outp-eval2-mode_thrFactor-pictureset15-cb.csv}};
            \addplot+ table 
                [x=accPixel, y=throughput, col sep=comma]
                {\plotdataDirPfm{outp-eval2-mode_thrFactor-pictureset16-cb.csv}};
                
            \draw[thin] (axis cs:\pgfkeysvalueof{/pgfplots/xmin},\baselineVal) -- (axis cs:\pgfkeysvalueof{/pgfplots/xmax},\baselineVal)
            node[above] at (axis cs:{\pgfkeysvalueof{/pgfplots/xmax}-1},\baselineVal) {cuDNN};
    	\end{axis}
    \end{tikzpicture}
    \caption{Evaluation of the throughput-accuracy trade-off for all 6 video sequences.}
    \label{fig:accVSperfAll}
\end{figure}
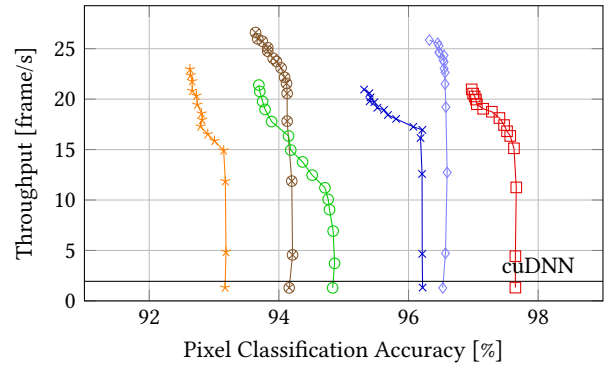
While for some scenarios any drop in accuracy is unacceptable, many applications allow for some trade-off between accuracy and throughput---after all choosing a specific CNN already implies selecting a network with an associated accuracy and computational cost. 

We analyze the trade-off directly in \figref{fig:accVSperfAll}. The most extreme case is updating the entire frame every time resulting in the lowest throughput at the same accuracy as full-frame inference. Increasing the threshold factor in steps of 0.25 immediately results in a significant throughput gain and for most sequences the trade-off only starts at frame rates close to saturation above \pfmVal{200}{16}\,frame/s. The same frame sequence already deviate from the norm before behaves differently here as well. However, an adaptive selection of the threshold factor such as a simple control loop getting feedback about the number of changed pixels could allow for a guaranteed throughput by reducing the accuracy in such cases and is left to be explored in future work.

\subsection{Compute Time Breakdown}
\begin{figure}
    \centering
    \scriptsize
    \begin{tikzpicture} 
        \begin{axis}[
                xbar stacked,
                symbolic y coords={L3 Conv., L2 Conv., L1 Conv.},
                height=3.2cm, width=\linewidth, xlabel={Compute Time [$\mu s$]},
                legend style={at={(0.5,-0.40)}, anchor=north,legend columns=-1}, enlargelimits=0.3, xmin=0, xmax=\pfmVal{850}{18900}, enlarge x limits=false,
                try min ticks=9, 
                ytick=data,
                ] 
            \addplot coordinates {( \pfmVal{89}{1990},L1 Conv.) ( \pfmVal{110}{4490},L2 Conv.) ( \pfmVal{127}{2819},L3 Conv.)}; 
            \addplot coordinates {( \pfmVal{134}{3102},L1 Conv.) (  \pfmVal{51}{1465},L2 Conv.) (  \pfmVal{46}{3016},L3 Conv.)}; 
            \addplot coordinates {( \pfmVal{76}{2069},L1 Conv.) ( \pfmVal{42}{1773},L2 Conv.) (\pfmVal{87}{1016},L3 Conv.)}; 
            \addplot coordinates {(  \pfmVal{478}{9341},L1 Conv.) ( \pfmVal{122}{2256},L2 Conv.) ( \pfmVal{477}{5490},L3 Conv.)}; 
            \addplot coordinates {(  \pfmVal{22}{374},L1 Conv.) (  \pfmVal{7}{285},L2 Conv.) (  \pfmVal{15}{187},L3 Conv.)}; 
            
            \legend{Change Det., Change Extr., gen. X, GEMM, Output Upd.}

        \end{axis} 
    \end{tikzpicture}
    \caption{Compute time for the individual processing steps per layer running on the GPU for a typical frame sequence.}
    \label{fig:timebreakdown}
\end{figure}
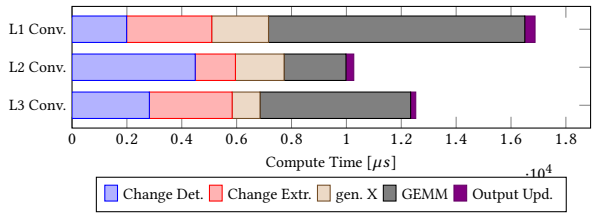
In Section~\ref{sec:baselineComputeBreakdown} and specifically in \tblref{tbl:baselineComputeTime}, we already discussed the compute time breakdown of the entire network when using frame-by-frame analysis. To gain more in-depth understanding of the limiting factors of our proposed algorithm, we show a detailed compute time breakdown of only the change-based convolution layers in \figref{fig:timebreakdown}. The time spent on change detection is similar across all 3 conv layers, which aligns well with our expectations since the feature map volume at the input of $n_{ch}\cdot h\cdot w$ values is identical for L2 and L3, and 25\% smaller for L1. That this step already makes up for more than \pfmVal{10}{23.4}\% of the overall time underlines the importance of a very simple change detection function: any increase in compute time for change detection has to be offset by time savings in the other steps by reducing the number of changes significantly. The change indexes extraction effort is linear to the number of pixels $h\cdot w$ and the clear drop from L1 to L2 is as expected. However, since it is not well parallelizable, there is not much additional gain when comparing L2 to L3. The effort to generate the $\mx{X}$ matrix is very dependent on the number of changed pixels, the number of feature maps, and the filter size. It is however mostly important that the time spent on shuffling data around to generate $\mx{X}$ is significantly smaller than the actual matrix multiplication, which clearly makes up the largest share. The subsequent update of the output values including activation only uses a negligible part of the overall processing time. 

An important aspect is not directly visible: The overall compute time for the critical part, the convolution and activation of L1-L3, has shrunk tremendously by more than $\pfmVal{15}{12.9}\times$ from \pfmVal{\us{29170}}{\ms{512.8}} to about \pfmVal{\us{1900}}{\ms{39.7}}. The remaining steps like polling (total \pfmVal{\us{600}}{\ms{6.6}}) and pixel-wise classification (L4, L5; total \pfmVal{\us{910}}{\ms{16.0}}) now take \pfmVal{44}{36}\% of the compute time, such that they move more into the target of future optimizations. 

\subsection{Change Propagation}
During the construction of the algorithm we argued that change detection should be performed for every convolution layer not only for modularity, but also justifying that the worst-case change propagation we had to assume otherwise would result in a higher computational effort. 
\begin{figure}
    \centering
	\includegraphics[width=\linewidth]{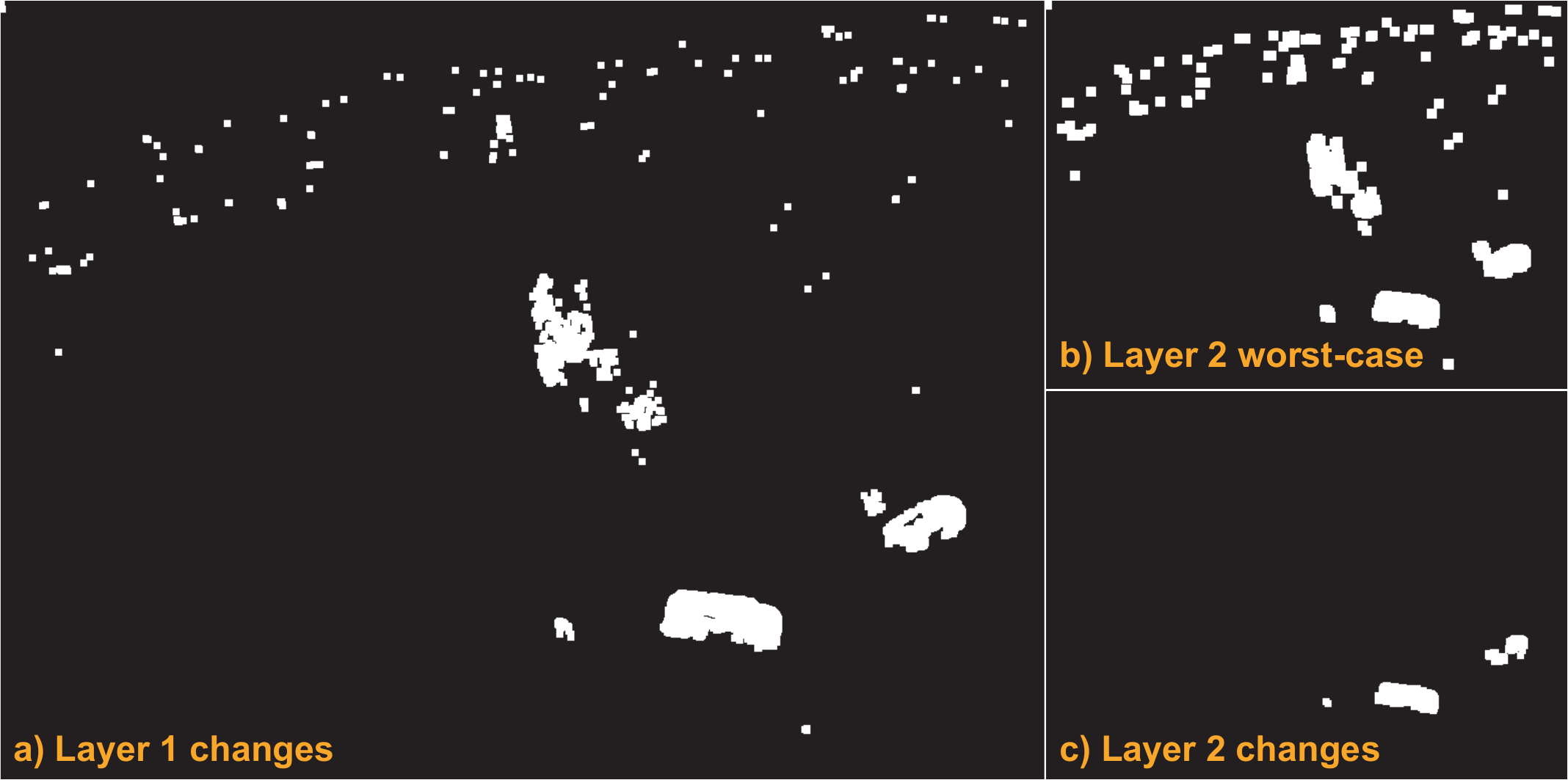}
    \caption{Analysis of the change propagation. (a) shows the changes detected in Layer~1 using the thresholds determined in Section~\ref{sec:ThresholdSelResults}, in the upper part of the image there are several single-pixel changes due to noise. We show the changed pixels for Layer~2 based on worst-case propagation as assumed when dropping the Layer~2 change detection step (b) and those when applying change detection instead (c).}
    \label{fig:changeProp}
\end{figure}
We have experimentally verified this and show an example case in \figref{fig:changeProp}. For Layer~2, the number of changes is reduced by $6.8\times$ from 7.57\% to 1.11\% and for Layer~3 from 2.58\% to 1.94\% by $1.33\times$. While it clearly pays of for Layer~2, we analyze the situation for Layer~3 more closely. From the previous section, we know that the change detection makes up for only \pfmVal{16}{22}\% of the overall compute time in Layer~3 and scaling up the time to generate the $\mx{X}$ matrix, perform the matrix multiplication and update the output clearly exceeds the overhead introduced by the change detection step. The change extraction step cannot be dropped, and in fact the change detection has to be replaced with a (though very quick to evaluate) change propagation kernel. 

\subsection{Energy Efficiency}
We have measured the current consumption of the entire Jetson TX1 board with only an Ethernet connection and no other off-board peripherals, running a continuous CNN workload. The average current when running the baseline cuDNN-based implementation was measured at \SI{680}{\milli\ampere} (\SI{12.9}{\watt}). The power consumption dropped to \SI{10.92}{\watt} when running CBinfer. When idling, we measured \SI{1.7}{\watt} under normal conditions, which raised to \SI{2.5}{\watt} when enforcing the maximum clock frequency, as has been done to maximize throughput for the earlier measurements. The CNN we used has a computational complexity of 210\,GOp/frame, where the number of operations (Ops) is the sum of additions and multiplications required for the convolution layers. We thus obtain 413\,GOp/s and 32.0\,GOp/s/W with the cuDNN baseline. With the proposed CBinfer procedure, we obtain a per-frame inference equivalent throughput of 3577\,GOp/s and an energy efficiency boost of about $10\times$ to 327.6\,GOp/s/W.

\section{Conclusion} \label{sec:conclusion}
We have proposed and evaluated a novel algorithm for change-based evaluation of CNNs for video recorded with a static camera setting, exploiting the spatio-temporal sparsity of pixel changes. The results clearly show that even when choosing the change detection parameters conservatively to introduce no significant increase in misclassified pixels during semantic segmentation, an average speed-up of $\pfmVal{7.6}{8.6}\times$ over a cuDNN baseline has been achieved using an optimized GPU implementation. An in-depth evaluation of the throughput-accuracy trade-off shows the aforementioned performance jump without loss and shows how the throughput can be further increased at the expense of accuracy. Analysis of the compute time split-up of the individual steps of the algorithm show that despite some overhead the GPU is fully loaded performing multiply-accumulate operations to update the changed pixels using the highly optimized cuBLAS matrix multiplication. An analysis of how changes propagate through the CNN further underline the optimality of the structure of the proposed algorithm. The resulting boost in energy efficiency over per-frame evaluation is an average of $10\times$, equivalent to 328\,GOp/s/W on the Tegra~X1 platform.

\section*{Acknowledgments}
\ifanonymized
\else
The authors would like to thank \emph{armasuisse Science \& Technology} for funding this research. This project was supported in part by the EU's H2020 programme under grant no. 732631 (OPRECOMP).
\fi

\bibliographystyle{acm-format-lukas}
\bibliography{IEEEabrv,mendeley}


\begin{thebibliography}{00}


\ifx \showCODEN    \undefined \def \showCODEN     #1{\unskip}     \fi
\ifx \showDOI      \undefined \def \showDOI       #1{{\tt DOI:}\penalty0{#1}\ }
  \fi
\ifx \showISBNx    \undefined \def \showISBNx     #1{\unskip}     \fi
\ifx \showISBNxiii \undefined \def \showISBNxiii  #1{\unskip}     \fi
\ifx \showISSN     \undefined \def \showISSN      #1{\unskip}     \fi
\ifx \showLCCN     \undefined \def \showLCCN      #1{\unskip}     \fi
\ifx \shownote     \undefined \def \shownote      #1{#1}          \fi
\ifx \showarticletitle \undefined \def \showarticletitle #1{#1}   \fi
\ifx \showURL      \undefined \def \showURL       {\relax}        \fi
\providecommand\bibfield[2]{#2}
\providecommand\bibinfo[2]{#2}
\providecommand\natexlab[1]{#1}
\providecommand\showeprint[2][]{arXiv:#2}

\bibitem[\protect\citeauthoryear{Abu-El-Haija, Kothari, Lee, Natsev, Toderici,
  Varadarajan, and Vijayanarasimhan}{Abu-El-Haija et~al\mbox{.}}{2016}]%
        {Abu-El-Haija2016}
\bibfield{author}{\bibinfo{person}{Sami Abu-El-Haija}, \bibinfo{person}{Nisarg
  Kothari}, {and} others.} \bibinfo{year}{2016}\natexlab{}.
\newblock \showarticletitle{{YouTube-8M: A Large-Scale Video Classification
  Benchmark}}.
\newblock  (\bibinfo{year}{2016}).
\newblock


\bibitem[\protect\citeauthoryear{Aimar, Mostafa, Calabrese, Rios-Navarro,
  Tapiador-Morales, Lungu, Milde, Corradi, Linares-Barranco, Indiveri, Liu, and
  Delbruck}{Aimar et~al\mbox{.}}{2017}]%
        {Aimar2017}
\bibfield{author}{\bibinfo{person}{Alessandro Aimar}, \bibinfo{person}{Hesham
  Mostafa}, {and} others.} \bibinfo{year}{2017}\natexlab{}.
\newblock \showarticletitle{{NullHop: A Flexible Convolutional Neural Network
  Accelerator Based on Sparse Representations of Feature Maps}}.
\newblock \bibinfo{journal}{{\em IEEE Transactions on Very Large Scale
  Integration Systems\/}} (\bibinfo{year}{2017}).
\newblock


\bibitem[\protect\citeauthoryear{Andri, Cavigelli, Rossi, and Benini}{Andri
  et~al\mbox{.}}{2016}]%
        {Andri2016}
\bibfield{author}{\bibinfo{person}{Renzo Andri}, \bibinfo{person}{Lukas
  Cavigelli}, {and} others.} \bibinfo{year}{2016}\natexlab{}.
\newblock \showarticletitle{{YodaNN: An Ultra-Low Power Convolutional Neural
  Network Accelerator Based on Binary Weights}}. In \bibinfo{booktitle}{{\em
  Proc. IEEE ISVLSI}}. \bibinfo{pages}{236--241}.
\newblock
\showISBNx{978-1-4673-9039-2}
\showDOI{%
\url{https://doi.org/10.1109/ISVLSI.2016.111}}


\bibitem[\protect\citeauthoryear{Cavigelli and Benini}{Cavigelli and
  Benini}{2016}]%
        {Cavigelli2016}
\bibfield{author}{\bibinfo{person}{Lukas Cavigelli} {and} \bibinfo{person}{Luca
  Benini}.} \bibinfo{year}{2016}\natexlab{}.
\newblock \showarticletitle{{Origami: A 803 GOp/s/W Convolutional Network
  Accelerator}}.
\newblock \bibinfo{journal}{{\em IEEE TCSVT\/}} (\bibinfo{year}{2016}).
\newblock
\showISBNx{9781450334747}
\showDOI{%
\url{https://doi.org/10.1145/2742060.2743766}}


\bibitem[\protect\citeauthoryear{Cavigelli, Bernath, Magno, and
  Benini}{Cavigelli et~al\mbox{.}}{2016}]%
        {Cavigelli2016a}
\bibfield{author}{\bibinfo{person}{Lukas Cavigelli}, \bibinfo{person}{Dominic
  Bernath}, {and} others.} \bibinfo{year}{2016}\natexlab{}.
\newblock \showarticletitle{{Computationally efficient target classification in
  multispectral image data with Deep Neural Networks}}. In
  \bibinfo{booktitle}{{\em Proc. SPIE Security + Defence}},
  Vol.~\bibinfo{volume}{9997}.
\newblock
\showDOI{%
\url{https://doi.org/10.1117/12.2241383}}


\bibitem[\protect\citeauthoryear{Cavigelli, Gschwend, Mayer, Willi, Muheim, and
  Benini}{Cavigelli et~al\mbox{.}}{2015a}]%
        {Cavigelli2015a}
\bibfield{author}{\bibinfo{person}{Lukas Cavigelli}, \bibinfo{person}{David
  Gschwend}, {and} others.} \bibinfo{year}{2015}\natexlab{a}.
\newblock \showarticletitle{{Origami: A Convolutional Network Accelerator}}. In
  \bibinfo{booktitle}{{\em Proc. ACM GLSVLSI}}. \bibinfo{publisher}{ACM Press},
  \bibinfo{pages}{199--204}.
\newblock
\showISBNx{9781450334747}
\showDOI{%
\url{https://doi.org/10.1145/2742060.2743766}}


\bibitem[\protect\citeauthoryear{Cavigelli, Magno, and Benini}{Cavigelli
  et~al\mbox{.}}{2015b}]%
        {Cavigelli2015}
\bibfield{author}{\bibinfo{person}{Lukas Cavigelli}, \bibinfo{person}{Michele
  Magno}, {and} \bibinfo{person}{Luca Benini}.}
  \bibinfo{year}{2015}\natexlab{b}.
\newblock \showarticletitle{{Accelerating Real-Time Embedded Scene Labeling
  with Convolutional Networks}}. In \bibinfo{booktitle}{{\em Proc. ACM/IEEE
  DAC}}.
\newblock


\bibitem[\protect\citeauthoryear{Chen, Krishna, Emer, and Sze}{Chen
  et~al\mbox{.}}{2016}]%
        {Chen2016}
\bibfield{author}{\bibinfo{person}{Yu-Hsin Chen}, \bibinfo{person}{Tushar
  Krishna}, {and} others.} \bibinfo{year}{2016}\natexlab{}.
\newblock \showarticletitle{{Eyeriss: An energy-efficient reconfigurable
  accelerator for deep convolutional neural networks}}. In
  \bibinfo{booktitle}{{\em Proc. IEEE ISSCC}}. \bibinfo{pages}{262--263}.
\newblock
\showISBNx{VO -}
\showISSN{01936530}
\showDOI{%
\url{https://doi.org/10.1109/ISSCC.2016.7418007}}


\bibitem[\protect\citeauthoryear{Chetlur, Woolley, Vandermersch, Cohen, Tran,
  Catanzaro, and Shelhamer}{Chetlur et~al\mbox{.}}{2014}]%
        {Chetlur2014}
\bibfield{author}{\bibinfo{person}{Sharan Chetlur}, \bibinfo{person}{Cliff
  Woolley}, {and} others.} \bibinfo{year}{2014}\natexlab{}.
\newblock \showarticletitle{{cuDNN: Efficient Primitives for Deep Learning}}.
\newblock


\bibitem[\protect\citeauthoryear{Collobert}{Collobert}{2011}]%
        {Collobert2011}
\bibfield{author}{\bibinfo{person}{Ronan Collobert}.}
  \bibinfo{year}{2011}\natexlab{}.
\newblock \showarticletitle{{Torch7: A Matlab-like Environment for Machine
  Learning}}.
\newblock \bibinfo{journal}{{\em Advances in Neural Information Processing
  Systems Workshops\/}} (\bibinfo{year}{2011}).
\newblock


\bibitem[\protect\citeauthoryear{Cordts, Omran, Ramos, Rehfeld, Enzweiler,
  Benenson, Franke, Roth, and Schiele}{Cordts et~al\mbox{.}}{2016}]%
        {Cordts2016}
\bibfield{author}{\bibinfo{person}{Marius Cordts}, \bibinfo{person}{Mohamed
  Omran}, {and} others.} \bibinfo{year}{2016}\natexlab{}.
\newblock \showarticletitle{{The Cityscapes Dataset for Semantic Urban Scene
  Understanding}}. In \bibinfo{booktitle}{{\em Proc. IEEE CVPR}}.
  \bibinfo{pages}{3213--3223}.
\newblock
\showISBNx{978-1-4673-8851-1}
\showISSN{10636919}
\showDOI{%
\url{https://doi.org/10.1109/CVPR.2016.350}}


\bibitem[\protect\citeauthoryear{Courbariaux, Bengio, and David}{Courbariaux
  et~al\mbox{.}}{2015}]%
        {Courbariaux2015a}
\bibfield{author}{\bibinfo{person}{Matthieu Courbariaux},
  \bibinfo{person}{Yoshua Bengio}, {and} \bibinfo{person}{Jean-Pierre David}.}
  \bibinfo{year}{2015}\natexlab{}.
\newblock \showarticletitle{{BinaryConnect: Training Deep Neural Networks with
  binary weights during propagations}}. In \bibinfo{booktitle}{{\em Adv.
  NIPS}}. \bibinfo{pages}{3105--3113}.
\newblock


\bibitem[\protect\citeauthoryear{Deng, Dong, Socher, Li, Li, and Fei-Fei}{Deng
  et~al\mbox{.}}{2009}]%
        {Deng2009}
\bibfield{author}{\bibinfo{person}{Jia Deng}, \bibinfo{person}{Wei Dong}, {and}
  others.} \bibinfo{year}{2009}\natexlab{}.
\newblock \showarticletitle{{ImageNet: A large-scale hierarchical image
  database}}. In \bibinfo{booktitle}{{\em Proc. IEEE CVPR}}.
\newblock
\showISBNx{978-1-4244-3992-8}
\showISSN{1063-6919}
\showDOI{%
\url{https://doi.org/10.1109/CVPR.2009.5206848}}


\bibitem[\protect\citeauthoryear{Farabet, Martini, Corda, Akselrod,
  Culurciello, and LeCun}{Farabet et~al\mbox{.}}{2011}]%
        {Farabet2011}
\bibfield{author}{\bibinfo{person}{Clement Farabet}, \bibinfo{person}{Berin
  Martini}, {and} others.} \bibinfo{year}{2011}\natexlab{}.
\newblock \showarticletitle{{NeuFlow: A Runtime Reconfigurable Dataflow
  Processor for Vision}}. In \bibinfo{booktitle}{{\em Proc. IEEE CVPRW}}.
  \bibinfo{pages}{109--116}.
\newblock
\showISBNx{978-1-4577-0529-8}
\showISSN{2160-7508}
\showDOI{%
\url{https://doi.org/10.1109/CVPRW.2011.5981829}}


\bibitem[\protect\citeauthoryear{Farrugia, Mamalet, Roux, {Fan Yang}, and
  Paindavoine}{Farrugia et~al\mbox{.}}{2009}]%
        {Farrugia2009}
\bibfield{author}{\bibinfo{person}{Nicolas Farrugia}, \bibinfo{person}{Franck
  Mamalet}, {and} others.} \bibinfo{year}{2009}\natexlab{}.
\newblock \showarticletitle{{Fast and Robust Face Detection on a Parallel
  Optimized Architecture Implemented on FPGA}}.
\newblock \bibinfo{journal}{{\em IEEE TCSVT\/}} \bibinfo{volume}{19},
  \bibinfo{number}{4} (\bibinfo{year}{2009}), \bibinfo{pages}{597--602}.
\newblock
\showISSN{1051-8215}
\showDOI{%
\url{https://doi.org/10.1109/TCSVT.2009.2014013}}


\bibitem[\protect\citeauthoryear{Fischer, Dosovitskiy, Ilg, Haeusser, Hazirbas,
  Glokov, Van~der Smagt, Cremers, and Brox}{Fischer et~al\mbox{.}}{2015}]%
        {Kaluarachchi2015}
\bibfield{author}{\bibinfo{person}{Philipp Fischer}, \bibinfo{person}{Alexey
  Dosovitskiy}, {and} others.} \bibinfo{year}{2015}\natexlab{}.
\newblock \showarticletitle{{FlowNet: Learning Optical Flow with Convolutional
  Networks}}. In \bibinfo{booktitle}{{\em arXiv:15047.06852}}.
\newblock


\bibitem[\protect\citeauthoryear{Gysel, Motamedi, and Ghiasi}{Gysel
  et~al\mbox{.}}{2016}]%
        {Gysel2016a}
\bibfield{author}{\bibinfo{person}{Philipp Gysel}, \bibinfo{person}{Mohammad
  Motamedi}, {and} \bibinfo{person}{Soheil Ghiasi}.}
  \bibinfo{year}{2016}\natexlab{}.
\newblock \showarticletitle{{Hardware-oriented Approximation of Convolutional
  Neural Networks}}. In \bibinfo{booktitle}{{\em ICLR Workshops}}.
\newblock
\showISBNx{9781369201741}


\bibitem[\protect\citeauthoryear{Han, Liu, Mao, Pu, Pedram, Horowitz, and
  Dally}{Han et~al\mbox{.}}{2016}]%
        {Han2016a}
\bibfield{author}{\bibinfo{person}{Song Han}, \bibinfo{person}{Xingyu Liu},
  {and} others.} \bibinfo{year}{2016}\natexlab{}.
\newblock \showarticletitle{{EIE: Efficient Inference Engine on Compressed Deep
  Neural Network}}.
\newblock \bibinfo{journal}{{\em arXiv:1602.01528\/}} (\bibinfo{year}{2016}).
\newblock


\bibitem[\protect\citeauthoryear{Hashemi, Anthony, Tann, Bahar, and
  Reda}{Hashemi et~al\mbox{.}}{2016}]%
        {Hashemi2016}
\bibfield{author}{\bibinfo{person}{Soheil Hashemi}, \bibinfo{person}{Nicholas
  Anthony}, {and} others.} \bibinfo{year}{2016}\natexlab{}.
\newblock \showarticletitle{{Understanding the Impact of Precision Quantization
  on the Accuracy and Energy of Neural Networks}}.
\newblock \bibinfo{journal}{{\em arXiv:1612.03940\/}} (\bibinfo{year}{2016}).
\newblock


\bibitem[\protect\citeauthoryear{He, Zhang, Ren, and Sun}{He
  et~al\mbox{.}}{2015a}]%
        {He2015}
\bibfield{author}{\bibinfo{person}{Kaiming He}, \bibinfo{person}{Xiangyu
  Zhang}, {and} others.} \bibinfo{year}{2015}\natexlab{a}.
\newblock \showarticletitle{{Deep Residual Learning for Image Recognition}}.
\newblock \bibinfo{journal}{{\em Proc. IEEE CVPR\/}} (\bibinfo{year}{2015}),
  \bibinfo{pages}{770--778}.
\newblock
\showISBNx{978-1-4673-8851-1}
\showDOI{%
\url{https://doi.org/10.1109/CVPR.2016.90}}


\bibitem[\protect\citeauthoryear{He, Zhang, Ren, and Sun}{He
  et~al\mbox{.}}{2015b}]%
        {HePReLU2015}
\bibfield{author}{\bibinfo{person}{Kaiming He}, \bibinfo{person}{Xiangyu
  Zhang}, {and} others.} \bibinfo{year}{2015}\natexlab{b}.
\newblock \showarticletitle{{Delving Deep into Rectifiers: Surpassing
  Human-Level Performance on ImageNet Classification}}.
\newblock


\bibitem[\protect\citeauthoryear{Held, Thrun, and Savarese}{Held
  et~al\mbox{.}}{2016}]%
        {Held2016}
\bibfield{author}{\bibinfo{person}{David Held}, \bibinfo{person}{Sebastian
  Thrun}, {and} \bibinfo{person}{Silvio Savarese}.}
  \bibinfo{year}{2016}\natexlab{}.
\newblock \showarticletitle{{Learning to track at 100 FPS with deep regression
  networks}}.
\newblock \bibinfo{journal}{{\em LNCS\/}}  \bibinfo{volume}{9905}
  (\bibinfo{year}{2016}), \bibinfo{pages}{749--765}.
\newblock
\showISBNx{9783319464473}
\showISSN{16113349}
\showDOI{%
\url{https://doi.org/10.1007/978-3-319-46448-0{\_}45}}


\bibitem[\protect\citeauthoryear{Iandola, Moskewicz, Ashraf, Han, Dally, and
  Keutzer}{Iandola et~al\mbox{.}}{2016}]%
        {Iandola2016}
\bibfield{author}{\bibinfo{person}{Forrest~N. Iandola},
  \bibinfo{person}{Matthew~W. Moskewicz}, {and} others.}
  \bibinfo{year}{2016}\natexlab{}.
\newblock \showarticletitle{{SqueezeNet: AlexNet-Level Accuracy with 50x Fewer
  Parameters and 1MB Model Size}}.
\newblock \bibinfo{journal}{{\em arXiv:1602.07360\/}} (\bibinfo{year}{2016}).
\newblock
\showISBNx{978-3-319-24552-2}
\showISSN{0302-9743}
\showDOI{%
\url{https://doi.org/10.1007/978-3-319-24553-9}}


\bibitem[\protect\citeauthoryear{Jaderberg, Vedaldi, and Zisserman}{Jaderberg
  et~al\mbox{.}}{2014}]%
        {Jaderberg2014a}
\bibfield{author}{\bibinfo{person}{Max Jaderberg}, \bibinfo{person}{Andrea
  Vedaldi}, {and} \bibinfo{person}{A Zisserman}.}
  \bibinfo{year}{2014}\natexlab{}.
\newblock \showarticletitle{{Speeding up Convolutional Neural Networks with Low
  Rank Expansions}}. In \bibinfo{booktitle}{{\em arXiv:1405.3866}}.
\newblock


\bibitem[\protect\citeauthoryear{Jia}{Jia}{2013}]%
        {Jia2013}
\bibfield{author}{\bibinfo{person}{Yangqing Jia}.}
  \bibinfo{year}{2013}\natexlab{}.
\newblock \bibinfo{title}{{Caffe: An Open Source Convolutional Architecture for
  Fast Feature Embedding}}.
\newblock   (\bibinfo{year}{2013}).
\newblock
\showURL{%
\url{http://caffe.berkeleyvision.org}}


\bibitem[\protect\citeauthoryear{Jin, Gokhale, Dundar, Krishnamurthy, Martini,
  and Culurciello}{Jin et~al\mbox{.}}{2014}]%
        {Jin2014}
\bibfield{author}{\bibinfo{person}{Jonghoon Jin}, \bibinfo{person}{Vinayak
  Gokhale}, {and} others.} \bibinfo{year}{2014}\natexlab{}.
\newblock \showarticletitle{{An efficient implementation of deep convolutional
  neural networks on a mobile coprocessor}}. In \bibinfo{booktitle}{{\em Proc.
  IEEE MWSCAS'14}}. \bibinfo{pages}{133--136}.
\newblock
\showISBNx{978-1-4799-4132-2}
\showDOI{%
\url{https://doi.org/10.1109/MWSCAS.2014.6908370}}


\bibitem[\protect\citeauthoryear{Karpathy, Toderici, Shetty, Leung, Sukthankar,
  and Li}{Karpathy et~al\mbox{.}}{2014}]%
        {Karpathy2014}
\bibfield{author}{\bibinfo{person}{Andrej Karpathy}, \bibinfo{person}{George
  Toderici}, {and} others.} \bibinfo{year}{2014}\natexlab{}.
\newblock \showarticletitle{{Large-scale video classification with
  convolutional neural networks}}.
\newblock \bibinfo{journal}{{\em Proc. IEEE CVPR\/}} (\bibinfo{year}{2014}),
  \bibinfo{pages}{1725--1732}.
\newblock
\showISBNx{9781479951178}
\showISSN{10636919}
\showDOI{%
\url{https://doi.org/10.1109/CVPR.2014.223}}


\bibitem[\protect\citeauthoryear{Krizhevsky, Sutskever, and Hinton}{Krizhevsky
  et~al\mbox{.}}{2012}]%
        {Krizhevsky2012a}
\bibfield{author}{\bibinfo{person}{Alex Krizhevsky}, \bibinfo{person}{Ilya
  Sutskever}, {and} \bibinfo{person}{Geoffrey~E Hinton}.}
  \bibinfo{year}{2012}\natexlab{}.
\newblock \showarticletitle{{Imagenet Classification With Deep Convolutional
  Neural Networks}}. In \bibinfo{booktitle}{{\em Adv. NIPS}}.
\newblock
\showISBNx{9781627480031}
\showISSN{10495258}


\bibitem[\protect\citeauthoryear{Kruegle}{Kruegle}{1995}]%
        {Kruegle1995}
\bibfield{author}{\bibinfo{person}{Herman Kruegle}.}
  \bibinfo{year}{1995}\natexlab{}.
\newblock \bibinfo{booktitle}{{\em {CCTV Surveillance: Video Practices and
  Technology}}}.
\newblock \bibinfo{publisher}{Butterworth-Heinemann}, \bibinfo{address}{Woburn,
  MA, USA}.
\newblock
\showISBNx{0-7506-9028-3}


\bibitem[\protect\citeauthoryear{Lavin}{Lavin}{2015}]%
        {Lavin2015}
\bibfield{author}{\bibinfo{person}{Andrew Lavin}.}
  \bibinfo{year}{2015}\natexlab{}.
\newblock \showarticletitle{{maxDNN: An Efficient Convolution Kernel for Deep
  Learning with Maxwell GPUs}}. In \bibinfo{booktitle}{{\em
  arXiv:1501.06633v3}}.
\newblock


\bibitem[\protect\citeauthoryear{Lavin and Gray}{Lavin and Gray}{2016}]%
        {Lavin2015a}
\bibfield{author}{\bibinfo{person}{Andrew Lavin} {and} \bibinfo{person}{Scott
  Gray}.} \bibinfo{year}{2016}\natexlab{}.
\newblock \showarticletitle{{Fast Algorithms for Convolutional Neural
  Networks}}. In \bibinfo{booktitle}{{\em Proc. IEEE CVPR}}.
  \bibinfo{pages}{4013--4021}.
\newblock
\showISBNx{978-1-4673-8851-1}
\showDOI{%
\url{https://doi.org/10.1109/CVPR.2016.435}}


\bibitem[\protect\citeauthoryear{Li, Kadav, Durdanovic, Samet, and Graf}{Li
  et~al\mbox{.}}{2016}]%
        {Li2016}
\bibfield{author}{\bibinfo{person}{Hao Li}, \bibinfo{person}{Asim Kadav}, {and}
  others.} \bibinfo{year}{2016}\natexlab{}.
\newblock \showarticletitle{{Pruning Filters for Efficient ConvNets}}.
\newblock \bibinfo{journal}{{\em arXiv:1608.08710\/}} (\bibinfo{year}{2016}).
\newblock


\bibitem[\protect\citeauthoryear{Lin, Maire, Belongie, Hays, Perona, Ramanan,
  Doll{\'{a}}r, and Zitnick}{Lin et~al\mbox{.}}{2014}]%
        {Lin2014}
\bibfield{author}{\bibinfo{person}{Tsung-Yi Lin}, \bibinfo{person}{Michael
  Maire}, {and} others.} \bibinfo{year}{2014}\natexlab{}.
\newblock \showarticletitle{{Microsoft COCO: Common Objects in Context}}. In
  \bibinfo{booktitle}{{\em Proc. ECCV}}.
\newblock
\showDOI{%
\url{https://doi.org/10.1007/978-3-319-10602-1{\_}48}}


\bibitem[\protect\citeauthoryear{Long, Shelhamer, and Darrell}{Long
  et~al\mbox{.}}{2015}]%
        {Long2015}
\bibfield{author}{\bibinfo{person}{Jonathan Long}, \bibinfo{person}{Evan
  Shelhamer}, {and} \bibinfo{person}{Trevor Darrell}.}
  \bibinfo{year}{2015}\natexlab{}.
\newblock \showarticletitle{{Fully Convolutional Networks for Semantic
  Segmentation}}. In \bibinfo{booktitle}{{\em Proc. IEEE CVPR}}.
\newblock


\bibitem[\protect\citeauthoryear{Park and Kim}{Park and Kim}{2016}]%
        {Park2016}
\bibfield{author}{\bibinfo{person}{Woon-sung Park} {and}
  \bibinfo{person}{Munchurl Kim}.} \bibinfo{year}{2016}\natexlab{}.
\newblock \showarticletitle{{CNN-based in-loop filtering for coding efficiency
  improvement}}. In \bibinfo{booktitle}{{\em Proc. IEEE Image, Video, and
  Multidimensional Signal Processing Workshop}}.
\newblock
\showISBNx{978-1-5090-1929-8}
\showDOI{%
\url{https://doi.org/10.1109/IVMSPW.2016.7528223}}


\bibitem[\protect\citeauthoryear{Paszke, Chaurasia, Kim, and
  Culurciello}{Paszke et~al\mbox{.}}{2016}]%
        {Paszke2016}
\bibfield{author}{\bibinfo{person}{Adam Paszke}, \bibinfo{person}{Abhishek
  Chaurasia}, {and} others.} \bibinfo{year}{2016}\natexlab{}.
\newblock \showarticletitle{{ENet: A Deep Neural Network Architecture for
  Real-Time Semantic Segmentation}}. In \bibinfo{booktitle}{{\em
  arXiv:1609.02147}}.
\newblock


\bibitem[\protect\citeauthoryear{Porikli, Bremond, Dockstader, Ferryman, Hoogs,
  Lovell, Pankanti, Rinner, Tu, and Venetianer}{Porikli et~al\mbox{.}}{2013}]%
        {Porikli2013}
\bibfield{author}{\bibinfo{person}{Fatih Porikli}, \bibinfo{person}{Francois
  Bremond}, {and} others.} \bibinfo{year}{2013}\natexlab{}.
\newblock \showarticletitle{{Video surveillance: past, present, and now the
  future [DSP Forum]}}.
\newblock \bibinfo{journal}{{\em IEEE Signal Processing Magazine\/}}
  \bibinfo{volume}{30} (\bibinfo{year}{2013}), \bibinfo{pages}{190--198}.
\newblock
\showISSN{1053-5888}
\showDOI{%
\url{https://doi.org/10.1109/MSP.2013.2241312}}


\bibitem[\protect\citeauthoryear{Rastegari, Ordonez, Redmon, and
  Farhadi}{Rastegari et~al\mbox{.}}{2016}]%
        {Rastegari2016}
\bibfield{author}{\bibinfo{person}{Mohammad Rastegari},
  \bibinfo{person}{Vicente Ordonez}, {and} others.}
  \bibinfo{year}{2016}\natexlab{}.
\newblock \showarticletitle{{XNOR-Net: ImageNet Classification Using Binary
  Convolutional Neural Networks}}. In \bibinfo{booktitle}{{\em
  arXiv:1603.05279}}.
\newblock


\bibitem[\protect\citeauthoryear{Redmon, Divvala, Girshick, and Farhadi}{Redmon
  et~al\mbox{.}}{2016}]%
        {Redmon2016}
\bibfield{author}{\bibinfo{person}{Joseph Redmon}, \bibinfo{person}{Santosh
  Divvala}, {and} others.} \bibinfo{year}{2016}\natexlab{}.
\newblock \showarticletitle{{You Only Look Once: Unified, Real-Time Object
  Detection}}.
\newblock


\bibitem[\protect\citeauthoryear{Ren, He, Girshick, and Sun}{Ren
  et~al\mbox{.}}{2015}]%
        {Ren2015}
\bibfield{author}{\bibinfo{person}{Shaoqing Ren}, \bibinfo{person}{Kaiming He},
  {and} others.} \bibinfo{year}{2015}\natexlab{}.
\newblock \showarticletitle{{Faster R-CNN: Towards Real-Time Object Detection
  with Region Proposal Networks}}.
\newblock \bibinfo{journal}{{\em arXiv:1506.01497\/}} (\bibinfo{year}{2015}).
\newblock


\bibitem[\protect\citeauthoryear{Russakovsky, Deng, Su, Krause, Satheesh, Ma,
  Huang, Karpathy, Khosla, Bernstein, Berg, and Fei-Fei}{Russakovsky
  et~al\mbox{.}}{2015}]%
        {Russakovsky2014}
\bibfield{author}{\bibinfo{person}{Olga Russakovsky}, \bibinfo{person}{Jia
  Deng}, {and} others.} \bibinfo{year}{2015}\natexlab{}.
\newblock \showarticletitle{{ImageNet Large Scale Visual Recognition
  Challenge}}.
\newblock \bibinfo{journal}{{\em IJCV\/}} \bibinfo{volume}{115},
  \bibinfo{number}{3} (\bibinfo{year}{2015}), \bibinfo{pages}{211--252}.
\newblock
\showISSN{0920-5691}
\showDOI{%
\url{https://doi.org/10.1007/s11263-015-0816-y}}


\bibitem[\protect\citeauthoryear{Shelhamer, Rakelly, Hoffman, and
  Darrell}{Shelhamer et~al\mbox{.}}{2016}]%
        {Shelhamer2016}
\bibfield{author}{\bibinfo{person}{Evan Shelhamer}, \bibinfo{person}{Kate
  Rakelly}, {and} others.} \bibinfo{year}{2016}\natexlab{}.
\newblock \showarticletitle{{Clockwork Convnets for Video Semantic
  Segmentation}}.
\newblock \bibinfo{journal}{{\em arXiv:1608.03609\/}} (\bibinfo{year}{2016}).
\newblock
\showISBNx{978-3-319-46447-3}
\showISSN{0302-9743}
\showDOI{%
\url{https://doi.org/10.1007/978-3-319-46448-0}}


\bibitem[\protect\citeauthoryear{Szegedy, Liu, Jia, Sermanet, Reed, Anguelov,
  Erhan, Vanhoucke, and Rabinovich}{Szegedy et~al\mbox{.}}{2015}]%
        {Szegedy2014}
\bibfield{author}{\bibinfo{person}{Christian Szegedy}, \bibinfo{person}{Wei
  Liu}, {and} others.} \bibinfo{year}{2015}\natexlab{}.
\newblock \showarticletitle{{Going Deeper with Convolutions}}. In
  \bibinfo{booktitle}{{\em Proc. IEEE CVPR}}.
\newblock


\bibitem[\protect\citeauthoryear{Vasilache, Johnson, Mathieu, Chintala,
  Piantino, and LeCun}{Vasilache et~al\mbox{.}}{2014}]%
        {Vasilache2014}
\bibfield{author}{\bibinfo{person}{Nicolas Vasilache}, \bibinfo{person}{Jeff
  Johnson}, {and} others.} \bibinfo{year}{2014}\natexlab{}.
\newblock \showarticletitle{{Fast Convolutional Nets With fbfft: A GPU
  Performance Evaluation}}.
\newblock \bibinfo{journal}{{\em arXiv:1412.7580\/}} (\bibinfo{year}{2014}).
\newblock


\bibitem[\protect\citeauthoryear{Zhang, Fang, Zhou, Pan, and Cong}{Zhang
  et~al\mbox{.}}{2016a}]%
        {Zhang2016a}
\bibfield{author}{\bibinfo{person}{Chen Zhang}, \bibinfo{person}{Zhenman Fang},
  {and} others.} \bibinfo{year}{2016}\natexlab{a}.
\newblock \showarticletitle{{Caffeine: Towards Uniformed Representation and
  Acceleration for Deep Convolutional Neural Networks}}. In
  \bibinfo{booktitle}{{\em Proc. ACM ICCAD}}. \bibinfo{address}{New York, NY,
  USA}.
\newblock
\showISBNx{9781450344661}
\showDOI{%
\url{https://doi.org/10.1145/2966986.2967011}}


\bibitem[\protect\citeauthoryear{Zhang, Liu, Wu, and Yang}{Zhang
  et~al\mbox{.}}{2016b}]%
        {Zhang2015b}
\bibfield{author}{\bibinfo{person}{Kaihua Zhang}, \bibinfo{person}{Qingshan
  Liu}, {and} others.} \bibinfo{year}{2016}\natexlab{b}.
\newblock \showarticletitle{{Robust Visual Tracking via Convolutional Networks
  without Training}}.
\newblock \bibinfo{journal}{{\em IEEE TIP\/}} (\bibinfo{year}{2016}).
\newblock
\showISSN{1057-7149}
\showDOI{%
\url{https://doi.org/10.1109/TIP.2016.2531283}}


\bibitem[\protect\citeauthoryear{Zhang, Wang, Tian, Yuan, and Xu}{Zhang
  et~al\mbox{.}}{2015}]%
        {Zhang2015a}
\bibfield{author}{\bibinfo{person}{Qian Zhang}, \bibinfo{person}{Ting Wang},
  {and} others.} \bibinfo{year}{2015}\natexlab{}.
\newblock \showarticletitle{{ApproxANN: An Approximate Computing Framework for
  Artificial Neural Network}}. In \bibinfo{booktitle}{{\em Proc. IEEE DATE}}.
\newblock
\showISBNx{9783981537048}
\showDOI{%
\url{https://doi.org/10.7873/DATE.2015.0618}}


\end{thebibliography}

\end{document}